\documentclass[12pt]{article}
\usepackage[margin=25mm]{geometry}
\usepackage{natbib}
\usepackage{amsmath}
\usepackage{amssymb}
\usepackage{authblk}
\usepackage{subcaption}
\usepackage{mathtools}
\usepackage{booktabs}
\usepackage{pgfplots}
\usepackage{hyperref}
\usepgfplotslibrary{fillbetween}
\usepgfplotslibrary{statistics}
\usetikzlibrary{arrows.meta, bending, positioning}

\pgfplotsset{compat=1.17}

\immediate\write18{texcount -tex -sum  \jobname.tex > \jobname.wordcount.tex}

\providecommand{\keywords}[1]
{
  \small	
  \textbf{\textit{Keywords---}} #1
}

\title{Reframing Audience Expansion through the Lens of Probability Density Estimation}

\author[]{Claudio Carvalhaes}
\affil[]{\small {C\&D Scientific Consulting, Inc.}\\ {c.g.carvalhaes@gmail.com}}
\date{}

\begin{document}

\maketitle

\begin{abstract}%
Audience expansion has become an important element of prospective marketing, helping marketers create target audiences based on a mere representative sample of their current customer base. Within the realm of machine learning, a favored algorithm for scaling this sample into a broader audience hinges on a binary classification task, with class probability estimates playing a crucial role. In this paper, we review this technique and introduce a key change in how we choose training examples to ensure the quality of the generated audience. We present a simulation study based on the widely used MNIST dataset, where consistent high precision and recall values demonstrate our approach's ability to identify the most relevant users for an expanded audience. Our results are easily reproducible and a Python implementation is openly available on GitHub: \url{https://github.com/carvalhaes-ai/audience-expansion}.
\end{abstract}


\section{Introduction}
\label{sec:intro}

Audience expansion is a methodology developed by ad-serving platforms to help advertisers find the best-matched audiences for their ads without looking into audience specifics. The rationale is that if you advertise to people who are similar to ones who already like the product or service you want to sell, chances are the conversion rate will be  high. By leveraging this methodology advertisers can effortlessly reach their ideal leads by simply uploading a list of reference individuals, also known as a seed audience, to the platform. Then, the platform expands this seed to an audience of the desired size, typically resulting in a significant reduction in customer acquisition costs compared to other targeting strategies.

From a machine learning perspective, a sound strategy for expanding a seed audience is by framing the problem as a binary classification task~\citep{qu2014systems,shen2015,liu2016audience,ma2016,ma2016c}. Essentially, this involves creating a two-class labeled training set, consisting of seed users and non-seed users, and then training a probabilistic classifier, e.g., Logistic Regression~\citep{jiang2019comprehensive}, to distinguish between the two classes. But instead of generating class predictions, the goal is to estimate the conditional probability that a given user belongs to the positive class. This probability is used to prioritize users for the expanded audience. Thanks to the arsenal of tools available for handling data complexities in a supervised learning setting, this approach has been reported as the preferred choice in most real-world implementations~\citep[p.20]{doan2021generative}. 

Nevertheless, while audience expansion is rooted in the idea of scoring users based on their similarity or proximity to seed users, probabilistic classification focuses on estimating the likelihood of class membership~\citep{bishop2006pattern}. This distinction is important because high probability scores are primarily driven by the classifier's confidence in class assignments, with no direct consideration for proximity to seed users. For instance, in a univariate Logistic Regression model, the probability of belonging to the positive class monotonically increases as the distance from the decision threshold grows within the positive class subspace, regardless of how close or far the target data point is from the training instances~\citep{niculescu2005predicting}. This issue extends also to non-linear models. Consequently, an expanded audience made up of the highest scoring users is not guaranteed to bear the closest resemblance to the seed audience.

This paper focuses on a specific scenario in binary classification where the estimated probabilities accurately reflect proximity to seed users in feature space. The key lies in the choice of negative training examples. Rather than sampling these examples from the user base, we employ artificial examples uniformly distributed across the data domain. In this scenario, in order to minimize classification errors, the classifier needs to function as a type of density estimation device, favoring users in regions with a denser concentration of seed examples for higher probability scores. In other words, a user assigned a high probability score is likely to be near seed users, thus encapsulating the core concept of audience expansion.

Naturally, the inherent complexity of non-parametric density estimation in high-dimensional space makes it difficult to accurately discern patterns and relationships within the user base. This problem is confronted in a two-phased manner. As a first step, we reduce data dimensionality using neighborhood-preserving projections~\citep{van2008visualizing}, which help in retaining the intrinsic structure of the data while simplifying its representation. Then, within this more manageable precomputed embedding space, we focus our efforts on model training and data expansion.

The effectiveness of our approach is demonstrated through simulated experiments. To this end, we use the well-known MNIST dataset~\citep{lecun1998mnist}, which serves as a surrogate user base. In this case, the embedding technique offers a representation wherein images of the same digit are closely positioned within the feature space. Thus, by sampling a specific digit to compose the seed audience, we can evaluate whether the probability scores can accurately pinpoint the remaining examples. The results consistently show high precision and recall values across multiple configurations of a simulated seed audience, while also highlighting the inability of traditional two-class classification to identify the most relevant instances for an expanded audience.

% The remainder of this paper is organized as follows. Section~\ref{sec:related_work} discusses related work. Section~\ref{sec:preliminaries} presents the notation and a brief overview of the classification approach to audience expansion. Section~\ref{sec:classification} outlines our proposed approach. Section~\ref{sec:experiment} describes the experiment used for performance evaluation, and Section~\ref{sec:results} shows the results. Finally, Section~\ref{sec:conclusion} concludes the paper with final remarks and future directions.

\section{Related Work}
\label{sec:related_work}

Audience expansion is typically addressed using similarity measures~\citep{liu2016audience, ma2016,ma2016c,chakrabarti2008contextual} or as a two-class classification problem~\citep{qu2014systems,shen2015}. In the case of similarity measures, coefficients such as Jaccard similarity and cosine similarity~\citep{jiang2019comprehensive} are commonly applied to estimate the average similarity score between a target user and all the seed users. The expanded audience consists of users exihibiting the highest similarity scores. The underlying assumption is that users with similar feature vectors exhibit similar responses to advertising campaigns, which undoubtedly fits well with the core idea behind audience expansion. However, similarity measures are particularly challenged by the wide range of data complexities found in ad-serving systems, wherein learning models are better equipped to thrive~\citep{doan2021generative,liu2019real}. 

Compared to similarity measures, classification algorithms have the advantages of being optimized for prediction accuracy, they can be more interpretable,  can learn nonlinear data dependencies, deal with missing data, and generalize to unseen data~\citep{murphy2012machine}. But unlike a standard classification task like distinguishing between spam and non-spam emails, where both classes are  characterized by `crispy' training examples, in this case this only applies to the positive class. A couple of strategies are commonly used to choose negative training examples. One procedure is to sample them from non-converted users, such as users who saw an ad but didn't click on it~\citep{qu2014systems,qiu2021methods}, or users who clicked on an ad but did not install the advertised app~\citep{ma2016b}. This procedure, however, suffers from the so-called cold start problem that occurs when not enough training examples are available in a newly created campaign~\citep{ma2016,doan2019adversarial}. A simpler procedure to circumvent this problem is to randomly select users from the user base~\citep{dewet2019finding,liu2019real,zhu2021learning}. However, the overall validity of the approach is called into question due to the potential for selecting users who resemble seed users as negative training examples~\citep{liu2016audience,ma2016,doan2019adversarial,jiang2019comprehensive}.

The approach presented in this paper aligns with the concept of ``unsupervised learning as supervised learning'', as described by \citet[sec.14.2.4]{hastie2009}. In this framework, the classifier takes on a role akin to that of a probability density estimator, but with the advantage of leveraging supervised learning techniques in an unsupervised setting. 

A significant difference between our approach and other classification-based procedures for audience expansion lies in our handling of negative training examples. Instead of sampling these examples exclusively from the user base, we extend our scope to encompass instances that do not correspond to actual users. Moreover, all the computations are carried out after embedding the user base in a lower-dimensional space.

Finally, it is worth noting that audience expansion can be viewed as a type of  one-class classification problem (OCC)~\citep{tax2004support}. However, the primary focus in OCC is on establishing class membership associations, whereas in the context of audience expansion the goal is to infer the likelihood that unlabeled users are drawn from the same distribution as the seed audience. There is also a parallel with dataset expansion~\citep{maharana2022review}, with the key distinction being that audience expansion aims to pinpoint existing users in a user base resembling the seed data, whereas dataset expansion generates new samples mirroring the seed.

\section{Preliminaries}
\label{sec:preliminaries}

We begin by defining the feature space, denoted as $\mathcal{X}$, which is a subset of $\mathbb{R}^d$, where $d$ is the number of real-valued features. These features are essentially data points or variables that can be used to describe and distinguish one user from another within a platform's user base. Let $U=\{x_i\}_{i=1}^N$ be our user base, comprising $N$ users, each  associated with a feature vector in $\mathcal{X}$. A seed audience, denoted as $S$, refers to a specific subset of users in $U$ that is provided by the marketer for expansion. We assume that $S$ is  characterized by an underlying, albeit unknown, probability density function $f(x)$. 

To establish a classification framework, we consider two independent and identically distributed (i.i.d) samples of users, namely $\{x^0_i\}_{i=1}^{n_0}$ and $\{x^1_i\}_{i=1}^{n_1}$, with the following probability distributions:
\begin{equation}
\{x^0_i\}_{i=1}^{n_0} \stackrel{\mathrm{i.i.d}}{\sim} g(x) \quad \mathrm{and}\quad \{x^1_i\}_{i=1}^{n_1} \stackrel{\mathrm{i.i.d}}{\sim} f(x),
\label{eq:sampling}
\end{equation}
where $g(x)$ is a distribution chosen by the platform.  We also consider a label space  $\mathcal{Y} = \{0,1\}$ and assign $y=0$  to the sample $\{x^0_i\}_{i=1}^{n_0}$ and $y=1$  to the sample $\{x^1_i\}_{i=1}^{n_1}$ to obtain a labeled training set $\{(x_i,y_i)\}_{i=1}^{n_0 + n_1}$ in $\mathcal{X} \times \mathcal{Y}$ upon which a probabilistic classifier can learn to distinguish between the two classes. The parameters $n_0$ and $n_1$, where $n_0 + n_1 \ll N$, control class weight. An approximately equal value for $n_0$ and $n_1$ is the ideal choice to mitigate the challenges posed by class imbalance.

In a probabilistic setting, a binary classifier learns to predict a conditional probability distribution $p(y\mid x)$ over the class labels subject to
\begin{equation}
p(0\mid x) + p(1\mid x) = 1.
\label{eq:probs_sum_to_1}
\end{equation}
The estimated probability $p(1|x_i)$, which gives the likelihood of a user $x_i \in U$ being drawn from the density $f(x)$, provides a score to prioritize users for the expanded audience. For instance, suppose a seed audience $S \subset U$ is given and the goal is to obtain an expanded audience consisting of $m$ prospects, which is typically specified as a percentage of $N$~\citep{dewet2019finding}. Then, the trained classifier is applied to $U$ and the expanded audience is given by
\begin{equation}
    A = \underset{A^\prime \subset U \cap S,\, \vert A^\prime \vert = m}{\mathrm{argmax}} \sum_{x_i \in A^\prime} p(1\mid x_i).
    \label{eq:expanded_audience}
\end{equation}
In words, the expanded audience consists of the $m$ highest-scoring users in $U$, excluding the seed audience.

Although class prediction is not the end goal, an important advantage of the classification approach is the possibility to use techniques like cross-validation for model selection and hyperparameter optimization. In this case, given a user $x_i \in U$, the class predicted by the classifier follows from the maximum \emph{a posteriori} rule, i.e., it is the most probable one:
\begin{equation}
    \mathrm{class}(x_i) = \underset{y_i\in \mathcal{Y}}{\mathrm{argmax}}\,p(y_i\mid x_i).
    \label{eq:decision_rule}
\end{equation}
At this stage, a variety of performance measures can be employed to evaluate different aspects of model performance for a given seed audience~\citep{tharwat2020classification}. 

\section{Methods}
\label{sec:methods}

\subsection{Issues with the traditional approach}
The classification framework described above has an important pitfall. While it is true that the probability  $p(1 | x)$ tends to be large near seed users, its values elsewhere may be more nuanced to interpret. As per Bayes' theorem:
\begin{equation}
    p(1\mid x)= \frac{f(x)\,P(1)}{g(x)\,P(0) + f(x)\,P(1)}= \frac{n_1\,f(x)}{n_0\,g(x) + n_1\,f(x)},
    \label{eq:bayes}
\end{equation}
where $x\in {\cal X}$ and the approximation $P(1)/P(0) \approx n_1/n_0$ was employed in a intermediary step. According to this equation,  $p(1 | x)$ is influenced not just by the distribution of seed users, but also by the particular choice of negative training examples. Therefore, choosing appropriate negative examples is crucial to ensure that $p(1|x)$ is higher near seed users and lower elsewhere. Simply focusing on obtaining a high classification rate on an independent test set, such as a standard classification task, is insufficient to guarantee this outcome. 

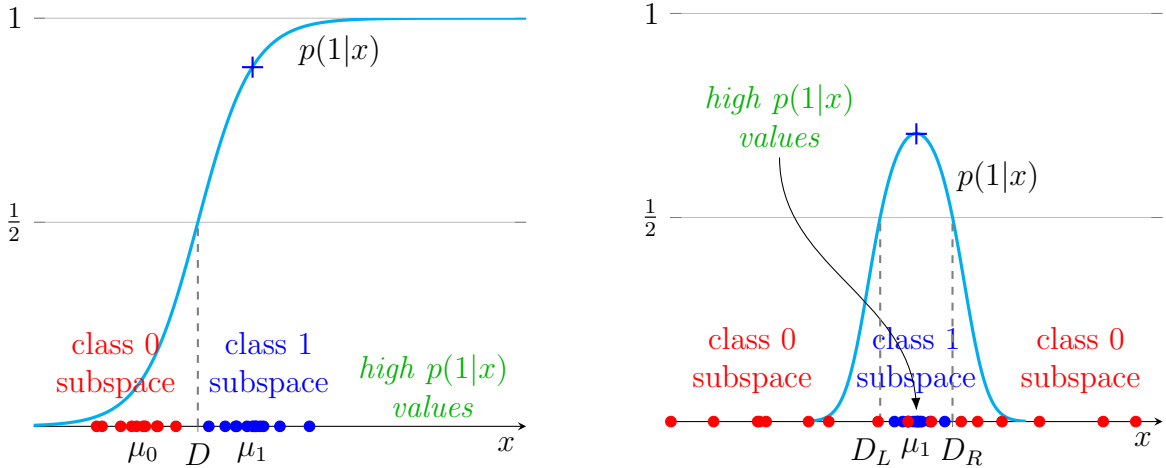
\begin{figure}
    \def\xl{1.678772588822513}
    \def\xr{4.321227411177487}
     \centering
    \begin{subfigure}{0.49\textwidth}
        \centering
        \begin{tikzpicture}
            \begin{axis}[
                xlabel=$x$,
                y axis line style={draw opacity=0},
                y axis line style=-,
                axis x line=middle,
                xlabel style = {at={(axis description cs:1,0)},anchor=north east},
                ylabel style = {at={(axis description cs:0,1)},anchor=south east},          
                xmin=-1, xmax=8,
                ymin=0, ymax=1,
                domain=-1:8,
                grid=both,
                xmajorgrids=false,
                samples=200,
                xticklabel style={font=\normalsize,fill=white},
                xtick={1, 2, 3},
                xticklabels={ $\mu_0$, $D$, $\mu_1$},
                yticklabel style={font=\normalsize,fill=white},
                ytick={1/2,1},
                yticklabels={$\frac{1}{2}$,  $1$},
                legend style={at={(0.0,0.8)},anchor=west,font=\small},
                width=\textwidth,
            ]
                \addplot[cyan, smooth, very thick, name path=f] {exp(2*x -4) / (1 + exp(2*x - 4)};
                \node[font=\normalsize] at (axis cs: 4.6, 0.92) {$p(1|x)$};
                \draw[gray, dashed, thick] (2,0) -- (2,0.5);
                \addplot[blue,mark=+,thick,mark size=4pt] coordinates {(3, 0.88)};
                \node[red,font=\normalsize,align=center] at (axis cs: 0.5, 0.15) {class 0\\ subspace};
                \node[blue,font=\normalsize,align=center] at (axis cs: 3.3, 0.15) {class 1\\ subspace};
                \node[black!30!green,font=\normalsize,align=center] at (axis cs: 6.3, 0.1) {\emph{high} $p(1|x)$\\ \emph{values}};
                \addplot[blue, only marks, mark=*, samples at={2.2,3.2,3.14,3.5,2.71,3.0,3.07,2.5,2.69,3.04,2.90,4.04}] {0};
                \addplot[red, only marks, mark=*, samples at={1.27,0.88 ,1.25,1.00,0.79,0.59,1.6,1.04,0.15,0.25}] {0};
            \end{axis}
        \end{tikzpicture}
    \end{subfigure}     
    \hfill
    \begin{subfigure}{0.49\textwidth}
        \centering
        \begin{tikzpicture}
            \begin{axis}[
                xlabel=$x$,
                % ylabel=$r(x)$,
                y axis line style={draw opacity=0},
                y axis line style=-,
                axis x line=middle,
                xlabel style = {at={(axis description cs:1,0)},anchor=north east},
                ylabel style = {at={(axis description cs:0,1)},anchor=south east},          
                xmin=-6, xmax=12,
                ymin=0, ymax=1,
                domain=-1:7,
                grid=both,
                xmajorgrids=false,
                samples=200,
                yticklabel style={font=\normalsize, fill=white},
                ytick={0.5,1},
                yticklabels={$\frac{1}{2}$, $1$},
                xticklabel style={font=\normalsize,fill=white},
                xtick={\xl, \xr, 3},
                xticklabels={$D_L\;\;{}$, $\;\;{}D_R$, $\mu_1$},
                width=\textwidth,
            ]
                \addplot[cyan, smooth, very thick, name path=f] {0.4 * exp((-(x - 3)^2 )/ 2) / (1.0/6 + 0.4 * exp((-(x - 3)^2 )/ 2))};
                \node[font=\normalsize] at (axis cs: 6, 0.6) {$p(1|x)$};
                \node[red,font=\normalsize, align=center] at (axis cs: -3, 0.15) {class 0\\ subspace};
                \node[blue,font=\normalsize, align=center] at (axis cs: 3, 0.15) {class 1\\ subspace};
                \node[red,font=\normalsize, align=center] at (axis cs: 9, 0.15) {class 0\\ subspace};
                \node[black!30!green,font=\normalsize, align=center](A) at (axis cs: -2, 0.75) {\emph{high} $p(1|x)$\\ \emph{values}};
                \node (B) at (3, 0) {};
                \draw[-{Latex[bend]}, out=270, in=90] (A.south) to (B.north);
                \draw[gray, dashed, thick] (\xl,0) -- (\xl,0.5);
                \draw[gray, dashed, thick] (\xr,0) -- (\xr,0.5);
                \addplot[blue,mark=+,thick,mark size=4pt] coordinates {(3, 0.705)};
                \addplot[blue, only marks, mark=*, samples at={2.2,3.2,3.14,3.5,2.71,3.0,3.07,2.5,2.69,3.04,2.90,4.04}] {0};
                \addplot[red, only marks, mark=*, samples at={-5.97, -4.41, -2.8, -2.73, -2.49, -0.94, -0.2, 1.6, 2.7, 3.55, 4.64, 5.24, 6.13, 7.5, 9.83, 11.02}] {0};
            \end{axis}
            \end{tikzpicture}
    \end{subfigure}
    \captionsetup{format=hang}
    \caption{Probability scores derived from Eq.~\eqref{eq:bayes} with two distinct configurations of negative training examples (shown in red), while maintaining the same seed sample (in blue) for both panels. In the left panel, the negative examples follow a normal distribution centered at \(x = \mu_0\), and in the right panel, they are uniformly distributed. The cross symbol marks the point \((\mu_1, p(1|\mu_1))\) for reference.}
    \label{fig:univariate_examples} 
\end{figure}

The left panel of Figure \ref{fig:univariate_examples} provides further insights into this problem. For illustrative purposes, we consider a simplified one-dimensional example in which the set $\{x^1_i\}_{i=1}^{n_1}$ is tightly clustered around a single point at $x=\mu_1$, following a normal distribution. In turn, the negative set $\{x^0_i\}_{i=1}^{n_0}$ was created from a normal distribution around a distant point $x=\mu_0$, aiming to portrait a group of seed counterexamples in a real-world context. As a result, the classification task can be easily addressed by any learning algorithm, eventually yielding a high classification rate. However, the main takeaway is that $p(1|x)$ is not a reliable indicator of how close a user is to seed users. In fact, despite Eq.\eqref{eq:bayes} indicating that $p(1|x)$ is high near the point $x=\mu_1$ — the location of the seed cluster — its values actually continue to increase steadily towards 1 as we move rightward, further away from any seed example.

\subsection{The proposed approach}

In our approach, the negative set $\{x^0_i\}_{i=1}^{n_0}$ is selected by uniformly sampling instances across the data domain, regardless of whether they belong to the user base or not. With this choice, where $g(x)$ is constant, it follows from Equation~\eqref{eq:bayes} that  $p(1 | x)$  depends monotonically on the density $f(x)$, thereby achieving its highest values where the seed population is more dense. Consequently, an expanded audience made up of the highest scoring users has a feature distribution similar to the seed audience.

The right panel of Figure \ref{fig:univariate_examples} illustrates this scenario. The set $\{x^1_i\}_{i=1}^{n_1}$ is the same as in the left panel, but the negative set $\{x^0_i\}_{i=1}^{n_0}$ spans uniformly over the data domain. Importantly, there's no requirement for this set to represent actual users from the user base, nor is there a need to exclude regions occupied by seed users. 

The classification task becomes significantly more challenging than in the left panel, potentially leading to inferior classification rates. In particular, Logistic Regression is no longer an appropriate classifier. Indeed, equation~\eqref{eq:bayes} establishes two decision thresholds, denoted as $D_L$ and $D_R$, highlighting the need for a non-linear classifier. Despite these complexities, $p(1 | x)$ proves to be an effective measure of similarity to seed users: it reaches its peak value precisely at the center of the seed cluster, at $x=\mu_1$, and rapidly decreases towards zero as we depart from the seed audience.

Therefore, our choice of negative training examples results in a probability score that can actually serve as a robust indicator of proximity to seed users.

\subsection{Dimensionality reduction}

Fundamentally, the described approach is a practical application of probability density estimation, achieved by means of probabilistic classification. However, in real-world scenarios, estimating a probability density without any prior knowledge of what we are looking for becomes increasingly challenging as the data dimensionality grows, usually requiring a multi-step approach~\citep{scott2015multivariate}. The root cause of this problem lies from the sparsity of data points in high-dimensional spaces, which makes it difficult to tell apart a genuine data distribution from other distributions that may fit the data equally well. This fundamental problem persists when using classification, although it may not be as readily apparent. Our strategy to account for it involves two key steps: first, we reduce the data dimensionality by embedding the data into a lower-dimensional space, and then we build the training set and expand the seed within the embedding space. The next Section demonstrates this procedure using a popular multivariate dataset.

\section{Results}
\label{sec:results}

\subsection{The data}
The MNIST dataset~\citep{lecun1998mnist} was employed to simulate a user base and demonstrate the effectiveness of our approach. Specifically, we tested its ability to identify the relevant instances of a target class when provided with a sample as a seed. 

The MNIST dataset consists of a collection of 70,000 grayscale images of handwritten digits ranging from 0 to 9, where each image is represented by 784 features shaped as a 28x28-pixel matrix. It was specifically chosen for this illustration due to its characteristics of high-dimensional data and widespread availability, ensuring the reproducibility of our results.  In this simulation, each pixel represents a user attribute and the digits simulate lookalike audiences to be identified by the classifier based on a seed sample.

\subsection{Embedding representation}
\begin{figure}
    \centering
    \includegraphics[width=0.4\linewidth]{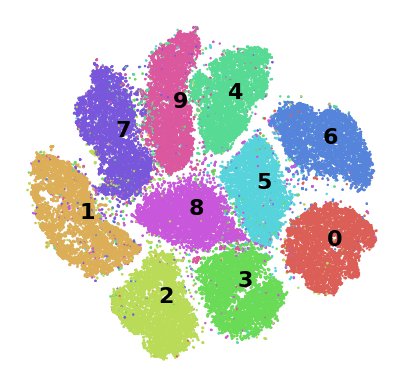}
    \captionsetup{format=hang}
    \caption{Embedding representation of the MNIST dataset.}
    \label{fig:embedding}
\end{figure}

The visualization technique of t-distributed stochastic neighbor embedding (t-SNE) \citep{van2008visualizing} was used to  transform the entire dataset into a two-dimensional map prior to any other tasks. A key feature of t-SNE is the preservation of the neighborhood structure of the data \citep{schubert2017intrinsic}, which is essential in audience expansion. The result is depicted in Figure~\ref{fig:embedding}. Note that because t-SNE lacks a direct functional mapping from the original space to the embedding space, it is not commonly used in model pipelines where test data is transformed using a function learned from training data. However, this does not affect our application as we are able to embed the entire population, i.e., the user base, before any computations take place.

\subsection{The learning task}
The learning task considered in this study consists of the following steps. Firstly, a specific digit from the embedded dataset is chosen as the true positive class, and 250 instances of the same digit are randomly selected as positive training examples. Negative training examples are artificially generated by uniformly sampling data points across the embedding space, resulting in a set of the same size as the positive examples. Secondly, a probabilistic classifier is trained on this data and employed to assign scores for belonging to the positive class, i.e., $p(1,x)$, to the remaining instances in the embedded dataset. This entire process is repeated 30 times for each of the ten digits, with a distinct training set being created in each repetition.

\subsection{Classification algorithm}
The Extra Trees algorithm ~\citep{geurts2006extremely} was chosen to perform the classification task. This choice was due to the algorithm's ability to capture complex patterns and interactions in the data, as well as its ease of interpretability. While a calibration technique was not required in the present illustration, it is important to acknowledge its potential significance to achieve a more reliable prediction model~\citep{platt1999probabilistic, zadrozny2002transforming, guo2017calibration, naeini2015obtaining}.

\subsection{Performance measure}
Performance evaluation was based on the precision and recall of the top-$k$ ranked instances by the classifier. These measures are commonly referred to as precision and recall at position $k$ and denoted by P@$k$ and R@$k$, respectively. In mathematical terms~\citep{liu2009learning},
\begin{equation}
    \mathrm{P}@k = \frac{\vert A_\mathrm{true} \cap A_k\vert}{\vert A_k \vert}\quad \mathrm{and} \quad \mathrm{R}@k = \frac{\vert A_\mathrm{true} \cap A_k\vert}{\vert A_\mathrm{true} \vert},
\end{equation}
where $A_\mathrm{true}$ is the set of true Class~1 instances in the pool (excluding the seed sample) and $A_k$ is the expanded audience of size $k$ as given by Eq.~\eqref{eq:expanded_audience}. In words, P@$k$ quantifies how many of the top-$k$ instances truly belong to Class~1, while R@$k$ represents the proportion of truly Class~1 instances that are successfully scored among the top-$k$ instances. The value of $k$ used in this experiment corresponds to 7,000 instances, which is approximately the total number of instances in each class.

\subsection{Results}
\begin{figure}
    \centering
    \begin{subfigure}{0.4\linewidth}
        \includegraphics[width=\linewidth]{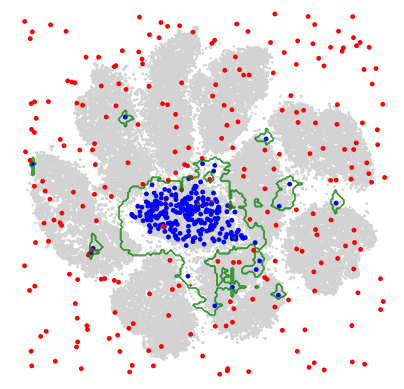}
    \end{subfigure}
    \hfill
    \begin{subfigure}{0.4\linewidth}
        \includegraphics[width=\linewidth]{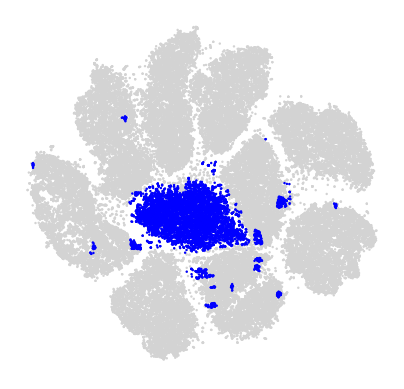}
    \end{subfigure}
    \captionsetup{format=hang}
    \caption{Left panel: A training set containing 250 examples from each class, along with the decision boundary learned by the classifier. Negative examples are represented by red dots, while positive examples are indicated by blue dots. Right panel: The resulting expanded audience made up of the top-ranked 7,000 instances in the embedding space.}
    \label{fig:boundary_and_scores}
\end{figure}

Figure~\ref{fig:boundary_and_scores} shows a training set example and the corresponding expanded audience generated by the density estimation approach in the t-SNE space. Observe that, in order to minimize classification errors, the classifier is forced to consider the subtleties of the seed distribution to carve out a decision boundary that isolates seed users (and the like) as much as possible from the negative examples. As a consequence, the probability score is consistently higher in the central region of the main cluster representing the seed class. 

\begin{figure}
    \centering
    \begin{subfigure}{0.4\linewidth}
        \includegraphics[width=\linewidth]{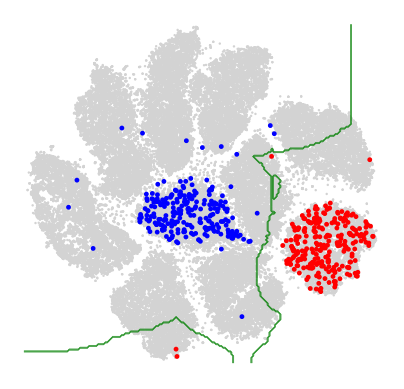}
    \end{subfigure}
    \hfill
    \begin{subfigure}{0.4\linewidth}
        \includegraphics[width=\linewidth]{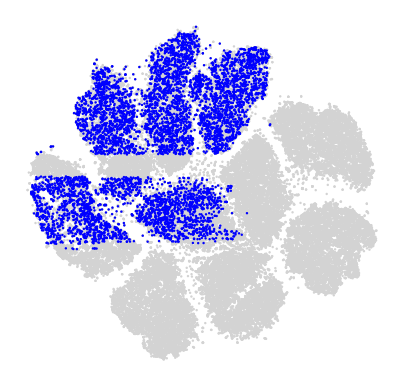}
    \end{subfigure}
    \captionsetup{format=hang}
    \caption{Illustration of a binary classification task distinguishing between digits 0 (red) and 8 (blue), with the latter designated as the positive class. The left panel displays the training samples alongside the decision boundary established by the classifier. The right panel shows the highest-ranked 7,000 instances selected from the user base. It is noteworthy that a majority of these instances fall outside the region associated with the true positive class.}
    \label{fig:6v8}
\end{figure}

For comparison, Figure~\ref{fig:6v8} illustrates the result of adopting a conventional classification task, where a second digit is chosen as the negative class. Note that the decision boundary is inherently less complex in this scenario than in Figure~\ref{fig:boundary_and_scores}. However, the ultimate measure of success is the quality of the expanded audience, which is visually of limited effectiveness, as it spreads to regions that do not correspond to the true positive class. The density estimation approach prevents this problem by learning a probability distribution that decays rapidly outside the region covered by the seed examples.

The results of the audience expansion experiment across all digits are summarized in Table~\ref{tab:patk}. Each entry in Table~\ref{tab:patk} represents the average value of each metric over the 30 repetitions of the expansion task. The average P@$k$ value across all digits was 0.90, indicating a high level of precision in identifying relevant instances in the expanded set. Similarly, the average R@$k$  value was 0.93, indicating a high level of recall in capturing relevant instances within the top-$k$ results. The consistent high values of P@$k$ and R@$k$ across all digits highlights the method's effectiveness in selecting the right instances for an expanded audience.
\begin{table}
    \centering
    \begin{tabular}[t]{ccc}
        \toprule
        Seed Class (Digit) & P@$k$ & R@$k$\\ 
        \midrule
         0 & 0.93 & 0.98\\
         1 & 0.96 & 0.88\\
         2 & 0.90 & 0.93\\
         3 & 0.89 & 0.91\\
         4 & 0.88 & 0.93\\
         5 & 0.82 & 0.95\\
         6 & 0.92 & 0.97\\
         7 & 0.92 & 0.91\\
         8 & 0.85 & 0.90\\
         9 & 0.88 & 0.91\\ 
        \bottomrule
    \end{tabular}
    \captionsetup{format=hang}
    \caption{Seed class representation and average performance values across 30 random repetitions of the expansion task. The overall average P@k value is 0.90, and the R@k value is 0.93.}
    \label{tab:patk}
\end{table}

\begin{figure}
    \centering
    \begin{subfigure}{0.4\linewidth}
        \includegraphics[width=\linewidth]{training-set-example.png}
    \end{subfigure}
    \hfill
    \begin{subfigure}{0.4\linewidth}
        \includegraphics[width=\linewidth]{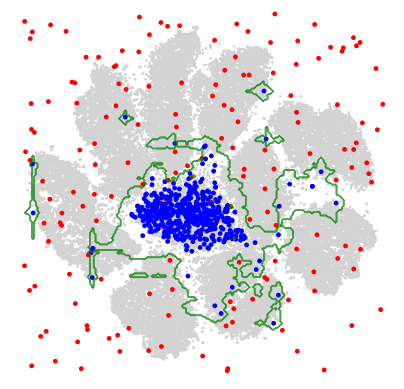}
    \end{subfigure}
    \captionsetup{format=hang}
    \caption{Decision boundary comparison between a balanced training set (left) and an unbalanced training set with an imbalance ratio of $n_1/n_0=3$ (right).}
    \label{fig:class_imbalance}
\end{figure}

It is worth noting that when expanding a tightly clustered seed, the expansion may need to occur primarily outside the decision boundary to encompass enough instances. However, the rapid decay of $p(1|x)$ outside a seed cluster makes it difficult to differentiate data points close to the boundary from those further away, ultimately impacting the quality of the expanded audience. A strategy to counter this issue in a controlled manner is to introduce class imbalance in favor of the positive class~\citep{ali2013classification}. By doing so, the decision boundary, defined by the condition $n_1\,f(x)=n_0\,g(x)$, is shifted away from the seed sample and towards a region where the density $f(x)$ is proportionally smaller than the uniform density $g(x)$ to compensate the imbalance factor. The resulting effect can be seen in Figure~\ref{fig:class_imbalance}, where the positive training examples outweigh the negative ones by a factor of three. Consequently, a larger region is allocated for expansion, while keeping focus on the seed distribution.

\section{Discussion}
\label{sec:discussion}

Previous studies by \citep{shen2015} and \citep{ma2016} have noted that audience expansion is a relatively underexplored research area, resulting in a scarcity of technical literature on the subject. In our current work, we aim to address this existing gap by introducing a key change to the classification approach for audience expansion. Unlike other classification-based methods, the primary focus is on generating probabilities that can serve as a measure of proximity to seed users, which is essential to ensure a high-quality expanded audience. The key steps involve: (1) generating a low-dimensional embedding representation of the user base that preserves the notion of neighborhood and (2) strategically selecting negative training examples that enable the classifier to generate probabilities that closely match the seed density.

Experimental results on the widely used MNIST dataset demonstrate the effectiveness of our approach, showing significantly high precision and recall values in identifying true instances of the positive class in a simulated user base. The choice of t-SNE as the embedding representation in this study proves effective in many scenarios. However, due to its local nature, it is important to acknowledge its potential suboptimal performance in situations with very high intrinsic dimensionality \citep{van2008visualizing}. Future research efforts could focus on strategies to overcome this limitation and further enhance the approach. 

\bibliography{main}

\end{document}